\documentclass[
]{ceurart}

\usepackage{listings}
\usepackage[inline]{enumitem}

\usepackage{tikz}
\usepackage{pgfplots}
\pgfplotsset{compat=1.18}
\usetikzlibrary{arrows.meta}
\usetikzlibrary{positioning,calc}

\usepackage{tabularx}
\usepackage{makecell}
\usepackage{todonotes}

\begin{document}

\copyrightyear{2026}
\copyrightclause{Copyright for this paper by its authors.
  Use permitted under Creative Commons License Attribution 4.0
  International (CC BY 4.0).}

\conference{R. Campos, A. Jorge, A. Jatowt, S. Bhatia, M. Litvak (eds.): Proceedings of the Text2Story'26 Workshop, Delft (The Netherlands), 29-March-2026}

\title{Hallucination or Creativity: How to Evaluate AI-Generated Scientific Stories?}

%
\author[1]{Alex Argese}[%
orcid=0009-0005-6151-5723,
email=alex.argese@eurecom.fr
]
\address[1]{EURECOM, Sophia Antipolis, France}

\author[1]{Pasquale Lisena}[%
orcid=0000-0003-3094-5585,
email=pasquale.lisena@eurecom.fr
]
\cormark[1]

\author[1]{Rapha\"{e}l Troncy}[%
orcid=0000-0003-0457-1436,
email=raphael.troncy@eurecom.fr
]

\cortext[1]{Corresponding author.}

\begin{abstract}
Generative AI can turn scientific articles into narratives for diverse audiences, but evaluating these stories remains challenging. Storytelling demands abstraction, simplification, and pedagogical creativity -- qualities that are not often well-captured by standard summarization metrics. Factual hallucinations are critical in scientific contexts, yet, detectors often misclassify legitimate narrative reformulations.
In this work, we propose StoryScore, a composite metric for evaluating AI-generated scientific stories. StoryScore integrates semantic alignment, lexical grounding, narrative control, structural fidelity, redundancy avoidance, and entity-level hallucination detection into a unified framework. Our analysis also reveals why many hallucination detection methods fail to distinguish pedagogical creativity from factual errors, highlighting a key limitation: while automatic metrics can effectively assess semantic similarity with original content, they struggle to evaluate how it is narrated and controlled.
\end{abstract}


\maketitle

\section{Introduction}
\label{sec:introduction}
Generative AI and Large language models (LLMs) can summarize and rephrase complex content, presenting it in a narrative form that makes it accessible to non-experts~\cite{teleki-etal-2025-survey}. Despite these advances, LLMs frequently struggle to appropriately adapt tone, level of detail, and stylistic choices to diverse audiences and communication objectives. In addition, LLMs are prone to hallucinations, eloquently stating unsupported claims, which pose a significant risk in scientific communication~\cite{ji2023survey}.

Recent works are exploring \emph{scientific storytelling} as a way to generate structured and engaging explanation of scientific content and papers, tailored to specific personas~\cite{roschelle2025generative,LI2023101311,sillano2026personas}. Evaluating such narratives presents challenges fundamentally different from those faced in traditional summarisation tasks. Storytelling deliberately introduces abstraction, metaphor, and contextualization to enhance accessibility. At the same time, this complicates evaluation: assessing factual grounding becomes non-trivial when the generated text is not expected to closely mirror the source document.
So, the question of how to evaluate generated narratives remains largely open. In particular:
\begin{enumerate}[label=\textbf{(RQ\arabic*)}]
    \item How can the quality of a scientific story be evaluated beyond surface-level similarity to the source paper, while accounting for narrative coherence and persona adaptation?
    \item How can hallucinations be reliably identified in stories where creative reformulation is not only expected but encouraged?
    \item To what extent do existing hallucination detection methods conflate legitimate abstraction with factual inconsistency in scientific storytelling?
\end{enumerate}

In this work, we address these questions in the context of designing a system that transforms scientific papers into audience-adapted stories. We make two main contributions:
\begin{enumerate}
    \item \textit{StoryScore}, a composite metric for automatic evaluation of scientific story generation;
    \item an empirical analysis of hallucination detection methods in persona-adaptive storytelling, showing why many approaches conflate creativity with hallucination.
\end{enumerate}

The metrics are grounded in a concrete use case: a two-stage pipeline (a \textit{Splitter} defining the outline and a \textit{Storyteller} in charge of the writing) that transforms scientific papers into persona-adapted stories.

The code used in this work is available at \url{https://github.com/AlexArgese/ai-scientist-storyteller}.

\section{Related Work}
\label{sec:related-work}
\paragraph{Narrative Storytelling in Science Communication.} Research shows that framing scientific content as a narrative significantly improves public comprehension, engagement~\cite{dahlstrom2014science_narrative} and recall of facts~\cite{freeman2024recall}. Scientific communication research emphasizes that effective dissemination requires deliberate choices about framing, analogy, structure, and level of detail, depending on the reader’s background and goals~\cite{national2017communicating,capili2024methods}. Recent work has also explored persona-driven or audience-aware generation to adapt explanations, tone, and terminology to reader profiles, showing benefits for accessibility and engagement but raising new requirements for evaluation beyond factuality alone \cite{roschelle2025generative,LI2023101311}.

In human–AI interaction settings, researchers are exploring tools to help scientists crafting such narratives. A recent example is \textit{RevTogether} that describes a multi-agent AI system that helps writers to revise “science stories” by blending engaging narrative structure with clear scientific content~\cite{zhang2025revtogether}.

\paragraph{Automatic evaluation of generated narratives.} Evaluating the quality of AI-generated summaries has also advanced beyond simple word-overlap scores. Traditional metrics like ROUGE~\cite{lin-2004-rouge} measure n-gram overlap with reference summaries, but they correlate poorly with human judgments~\cite{afzal2023challenges,scialom-etal-2021-questeval,chhun-etal-2022-human}. Semantic metrics such as BERTScore~\cite{zhang2019bertscore} and MoverScore~\cite{zhao-etal-2019-moverscore}  use contextual embeddings to better capture semantic alignment and compare meaning rather than surface words. They achieve high correlation with human ratings on summarization, but remain largely insensitive to structural degradation, redundancy, and discourse-level control failures. This gap is amplified in multi-section stories, where discourse-level issues (e.g., global coherence or instruction leakage) strongly affect perceived quality while remaining weakly captured by similarity-based metrics. Other methods, such as QuestEval~\cite{scialom-etal-2021-questeval} and QAGS~\cite{wang-etal-2020-asking}, ask questions about a summary and its source to detect inconsistencies.

\paragraph{Hallucination detection.} Hallucination detection has been studied extensively in summarization and question answering, using entity-based checks, retrieval and entailment methods, and LLM-based judges~\cite{ji2023survey,huang2025survey,li2025generation}. Systems can use RAG-based techniques on specific documents or on the web, like the widely adopted GPTZero\footnote{\url{https://gptzero.me/}}. Recent empirical work shows pitfalls in automated detection. For example, they show that ROUGE and similar metrics can be easily fooled -- e.g. repetitively duplicating correct content (adding length) artificially inflates ROUGE scores, even though no new facts are added~\cite{janiak-etal-2025-illusion}. In addition, common approaches often assume strict fidelity to the source. In persona-adaptive storytelling, legitimate reformulations and contextual expansions are expected. Thus, detectors can over-penalize creativity or behave unstably. We will critically analyse existing methods in Section~\ref{sec:hallucination}.



\section{The \textit{StoryScore} metric for Generated Story Evaluation}
\label{sec:storyscore}
The evaluation of stories generated from research papers, and in general, generated from a source material, requires metrics that balance faithfulness, completeness, and narrative quality under conditions of intentional abstraction. An effective evaluation framework should ensure that a generated story
\begin{enumerate*}[label=(\roman*)]
    \item \textbf{maintains semantic fidelity and representativeness} with respect to the source work, capturing the key content accurately even under extensive paraphrasing or reorganization;
    \item \textbf{preserves textual integrity}, minimizing artifacts and avoiding hallucinations introduced during generation; and
    \item \textbf{achieves structural and communicative adequacy}, ensuring coherent organization, appropriate titling, and elimination of unnecessary repetition.
\end{enumerate*}
These requirements call for an evaluation approach that integrates complementary signals rather than relying on a single notion of similarity or correctness.

For this reason, we propose \textit{\textbf{StoryScore}}, a composite metric aggregating semantic similarity, lexical grounding, structural fidelity, fluency, and hallucination control into a single score in the range $[0,1]$. Following extensive experimentation on a real use case, it is defined as:

\begin{equation}
\begin{aligned}
\mathit{StoryScore} =\;&
(0.3\,\mathit{ContextRecall}) \;+\;
(0.2\,\mathit{BERTScore}) \;+\; 
(0.2\,\mathit{PromptCleanliness}) \;+\; \\
(&0.1\,\mathit{TitleCoverage}) \;+\; 
(0.1\,\mathit{NoRedundancy}) \;+\; 
(0.1\,\mathit{NoHallucination})
\end{aligned}
\end{equation}

This formulation captures the most stable and reliable indicators, maintaining a balanced representation of narrative quality dimensions. The weights were selected heuristically after exploratory adjustments guided by manual qualitative inspection, with the goal of producing an aggregate score. The components of \textit{StoryScore} are summarised in Table~\ref{tab:metrics}.

\begin{table}[tb]
\caption{Summary of evaluation metrics used in this work. All components are in the $[0,1]$ range.}
\label{tab:metrics}
\centering
\begin{tabular}{l l l}
\toprule
\textbf{Metric} & \textbf{Type} & \textbf{Objective} \\
\midrule

Context Recall & Lexical &
Proportion of tokens from the original paper covered by the story  \\

BERTScore & Semantic &
Faithfulness of the story to the original content  \\

Prompt Cleanliness & Structural &
Absence of instruction leakage and prompt-related artifacts  \\

Title Coverage & Structural &
Similarity between generated and original section titles  \\

NoRedundancy & Fluency &
Avoidance of repeated $n$-gram loops and redundant phrasing patterns  \\

NoHallucination & Factuality &
Entity consistency (PERSON/ORG as detected by SpaCy) with the paper  \\

\end{tabular}
\end{table}

\textbf{Context recall} or \textbf{Article coverage} quantifies the amount of content from the original article that is reflected in the generated story, measured as the proportion of word-level lexical tokens from the article that also appear in the narrative (article-centered coverage). Tokens are defined as lowercased words, excluding punctuation and stopwords. Unlike metrics that rely solely on lexical overlap, contextual recall serves as a proxy for content coverage.
Formally, it is defined as:
\begin{equation}
\mathit{ContextRecall} =
\frac{\lvert T_{\mathrm{story}} \cap T_{\mathrm{paper}} \rvert}
     {\lvert T_{\mathrm{paper}} \rvert}
\end{equation}
where $T_{\mathrm{paper}}$ and $T_{\mathrm{story}}$ denote the sets of tokens extracted from the paper and the generated story. Higher values suggest a stronger connection to the original source material but also a vocabulary more faithful to the paper's language and less suited for a less knowledgeable audience. We deliberately adopt a simple set-based formulation to achieve a transparent and stable signal. This choice prioritises lexical grounding over semantic abstraction and may therefore disadvantage aggressive simplification.

\textbf{BERTScore} \cite{zhang2019bertscore} measures \textbf{semantic faithfulness} by comparing contextual embeddings of the generated story with those of the source article. BERTScore aligns tokens from the hypothesis to the most semantically similar tokens in the reference using cosine similarity in the embedding space. For each token, the F1-score is computed by aggregating the highest-scoring matches. Formally, let $H$ and $R$ be the embedding sets for the hypothesis (story) and reference (paper), BERTScore is defined as:
\begin{equation}
\mathit{BERTScore} = F_1 \bigl( \mathit{sim}(H,R),\, \mathit{sim}(R,H) \bigr)
\end{equation}
where $\mathit{sim}(A,B)$ denotes the cosine-similarity-based alignment between tokens in $A$ and their best matches in $B$. Higher values indicate stronger semantic grounding and better preservation of meaning.

\textbf{Prompt Cleanliness} measures the extent to which the generated story is free of \textbf{prompt-related artifacts} and \textbf{instruction leaks}, such as residual system directives, role indicators, formatting constraints, or meta-level indications. These artifacts indicate a failure of narrative abstraction, where the model collapses from storytelling to instruction-following behaviour, severely degrading readability.
The generated text is analysed line by line and sentence by sentence using a set of regular expression patterns that we empirically found as relevant.
A contamination score $C$ is computed as:
\begin{equation}
\begin{aligned}
C = \frac{1.0 \cdot N_{\mathrm{line}} + 0.75 \cdot N_{\mathrm{sent}} + 1.25 \cdot N_{\mathrm{json}}  + 0.75 \cdot N_{\mathrm{fence}} + 2.5 \cdot N_{\mathrm{block}}}{|L|}
\end{aligned}
\end{equation}
where $L$ denotes the set of non-empty lines in the generated text, and:

\begin{itemize}
\setlength\itemsep{-0.2em}
 \item $N_{\mathrm{line}}$ is the number of lines beginning with explicit instruction markers (e.g., \texttt{Human:}, 
 \texttt{Rules:});
 \item $N_{\mathrm{sent}}$ counts sentences that exhibit imperative instruction patterns at sentence boundaries;
 \item $N_{\mathrm{json}}$ denotes the number of lines consisting solely of structured JSON-like content;
 \item $N_{\mathrm{fence}}$ counts occurrences of markdown code fences;
 \item $N_{\mathrm{block}}$ counts dense instruction blocks characterized by multiple repeated occurrences of imperative constraints (e.g., three or more instances of “do not” within a single sentence or paragraph).
\end{itemize}
The weights associated with each term are selected empirically to prioritize the detection of severe leakage patterns over recall, assigning stronger penalties to artifacts that indicate a near-complete collapse into instruction-following mode.
The score is clipped to the unit, as outputs exceeding this threshold are considered equally dominated by prompt artifacts. Prompt Cleanliness is then defined as:
\begin{equation}
\mathit{PromptCleanliness} = 1 - \min(1, C)
\end{equation}


\textbf{Title Coverage} evaluates whether the generated story preserves the \textbf{section structure} of the target outline\footnote{In our use case, this is the outline generated by the \textit{Splitter} module, as in Section~\ref{sec:introduction}.}. Let $\mathcal{O} = \{o_1,\dots,o_5\}$ denote the target section titles and $\mathcal{G} = \{g_1,\dots,g_5\}$ the generated ones. We compute a soft similarity score and average it across sections:
\begin{equation}
\mathit{TitleCoverage} =
\frac{1}{n}\sum_{i=1}^{n} \mathrm{sim}\bigl(\mathrm{norm}(g_i),\,\mathrm{norm}(o_i)\bigr)
\end{equation}
where $n$ is the number of sections, $\mathrm{norm}(\cdot)$ removes differences in case, whitespaces, and punctuation, and $\mathrm{sim}(\cdot,\cdot)\in[0,1]$ is a string similarity function (1 for identical titles, lower values for partial matches). This yields a graded measure of structural fidelity that is robust to minor formatting differences.

\textbf{No Redundancy} is a \textbf{fluency indicator} that penalizes degenerative loops and excessive reuse of the same textual fragments, which are common artifacts in long-form generation. Repetition is quantified by computing the frequency of word-level $n$-grams in the generated story, with $n=3$ (trigrams), which provide a robust balance between sensitivity to repetition and tolerance to natural phrasing. Let $\mathcal{G}_n$ denote the multi-set of all $n$-grams extracted from the narrative, the repetition rate is defined as:
\begin{equation}
\mathit{RedundancyRate} =
\frac{\max_{g \in \mathcal{G}_n} \mathrm{freq}(g)}
     {|\mathcal{G}_n|},
\qquad
\mathit{NoRedundancy} = 1 - \mathit{RedundancyRate}
\end{equation}
High values of $\mathit{NoRedundancy}$ indicate fluent, varied text with minimal looping or redundant phrasing. This formulation is designed to capture obvious degenerative loops rather than fine-grained stylistic repetition, prioritizing robustness and interpretability over sensitivity to subtle discourse patterns.

\textbf{No Hallucination} quantifies the inclusion of entities that are not supported by the source material. After a post-processing normalization step applied to extracted entities (e.g.\ lowercasing, removal of possessive suffixes), we perform a NER-based comparison: the generated story is analyzed using SpaCy~\cite{spacy2020} to extract entities of type PERSON and ORG, which are then matched against the entities detected in the source paper. The restriction to these entity types is deliberate: in our manual qualitative inspection of generated outputs, hallucinations most frequently manifested as invented author names and institutional affiliations, making PERSON/ORG the most relevant entity types for a stable automatic signal. Any entity appearing in the story but absent from the source material is treated as a hallucination. A broader discussion of alternative hallucination detectors is provided in Section~\ref{sec:hallucination}. Formally, let $\mathit{GeneratedEntities}$ be the set of PERSON/ORG entities extracted from the story, and $\mathit{HallucinatedEntities}$ the subset of those not found in the paper, the score is defined as:
\begin{equation}
\label{eqn:no-allucination}
\mathit{NoHallucination} =
1 -
\frac{\lvert \mathit{HallucinatedEntities} \rvert}
     {\lvert \mathit{GeneratedEntities} \rvert}
\end{equation}

\section{Hallucination Detection in Scientific Storytelling}
\label{sec:hallucination}

Unlike summarization, storytelling intentionally involves simplification, contextualization, and narrative adaptation, producing an expected and desirable creative divergence. The core challenge is in distinguishing \textit{legitimate narrative abstraction} from \textit{true factual hallucination}, a distinction that existing hallucination detection methods (largely designed for summarization) are not well equipped to make.

The following analysis is grounded in a qualitative inspection of AI-generated stories produced by the pipeline on a representative subset of papers from the corpus. Each hallucination detection method was applied to the same generated stories and its outputs were manually examined to identify systematic failure modes specific to persona-adaptive scientific storytelling.

In this setting, hallucination cannot be reduced to mere deviation from the source text: a story may remain faithful while employing metaphors, analogies, or contextualization absent from the original paper. This inherent ambiguity makes hallucination detection particularly challenging and motivates the comparison of multiple detection approaches.

\paragraph{Capitalised Words as Entity Proxies.} This approach is deliberately simplistic: any word starting with a capital letter is treated as a potential factual entity. If such a token appears in the story but not in the article, it is marked as hallucination. This heuristic reveals clear limitations:
\begin{itemize}
\setlength\itemsep{-0.2em}
    \item abbreviations in the story (e.g.\ ``AI'') are flagged if the source document only contained the expanded form (e.g.\ ``Artificial Intelligence''),
    \item metaphors or narrative constructs capitalised for emphasis are misclassified as entities,
    \item minor morphological variants (pluralisation, genitives) produce false positives.
\end{itemize}

Although rudimentary, this method lays the foundation for understanding the structure of the problem: hallucination detection must separate surface-form noise from true semantic divergence.

\paragraph{NER-Based Detection (SpaCy PERSON/ORG).} We leverage SpaCy to only detect named entities of types PERSON and ORG, as described in Section~\ref{sec:storyscore}. This significantly reduces the number of candidates and detects genuinely invented organisations or people introduced by the model. Nevertheless, NER remained a shallow signal: it can detect incorrect affiliations, but fails to identify deeper factual inconsistencies, e.g. wrong scientific claims, invented datasets, or unsupported causal statements. We specifically focus on PERSON/ORG because our qualitative analysis repeatedly found fabricated authors and affiliations to be the most frequent and disruptive hallucination pattern in our stories.

\paragraph{MIRAGE Rewrite-Consistency Scoring.} MIRAGE is a library for hallucination detection based on rewriting the same passage multiple times and measuring the stability of the appearing concepts~\cite{10.1145/3746252.3761644}. If an idea disappears or mutates across rewrites, MIRAGE treats it as hallucinated. Although effective in summarization, this approach does not transfer well to persona-oriented storytelling. In our experiments, MIRAGE consistently and incorrectly flags explanatory metaphors (e.g., introducing a system by analogy with a familiar real-world process that is not explicitly described in the paper) and rephrasing for non-expert audiences (e.g., replacing technical terminology with higher-level conceptual descriptions). This indicates a misalignment between MIRAGE’s operational definition of fidelity, centred on literal grounding, and the goals of narrative systems that intentionally simplify the source material.

\paragraph{LLM-as-a-Judge.} We use an LLM to judge hallucinations directly. The judge receives the full scientific article as \texttt{CONTEXT}, the story as \texttt{ANSWER}, and produces a structured JSON verdict that includes faithfulness, hallucinated entities and numerical errors. This brings a qualitative leap because the LLM can reason about paraphrases and understand the broader context.

A first experiment conducted with the Qwen2.5-7B model~\cite{qwen2025qwen25technicalreport} has shown two systematic patterns:
\begin{enumerate}
\setlength\itemsep{-0.2em}
    \item \textbf{Missed hallucinations:} Some fabricated facts (e.g. invented affiliations) are left undetected.
    \item \textbf{False positive:} In several cases, it labeled hallucinated entities that were explicitly supported by the source paper (and mentioned in the story), i.e., the judge itself hallucinated the hallucination.
\end{enumerate}
In short, the judge sometimes \emph{hallucinates hallucinations}, making it too unreliable.

A second experiment used the GPT~5.1 model~\cite{chatgpt5.1}. While its assessments were more consistent with human judgment, the model remained overly strict, labelling benign contextual expansions as hallucinations merely because the source material did not explicitly mention those cases.

\paragraph{Hybrid Hallucination Detection (HHD).} A collection of some positive results from previous discoveries are combined together to form a hybrid technique consisting to:
\begin{enumerate}
\setlength\itemsep{-0.2em}
    \item extract “technical tokens” via SpaCy (capitalised words, acronyms, numbers),
    \item retrieve the top-$k$ most similar sentences from the article using MiniLM embeddings~\cite{wang202minilm},
    \item mark a token as hallucinated only if it appears in none of the retrieved contexts \emph{and} the similarity score is below a threshold.
\end{enumerate}
This approach combines symbolic robustness (entity extraction), semantic flexibility (retrieval-based context), and local grounding (sentence-level comparison), but is difficult to calibrate.

Manual inspection on generated stories showed that false positives dominate the detector’s output, mainly due to pedagogical reformulations that are conceptually faithful but not literally supported by retrieved sentences. For example, the following excerpt is from a generated story about~\cite{gao2024training}:
\begin{quote}
\emph{\textbf{Hermes}, the messenger god of ancient Greece, was known for his speed and efficiency. Similarly, the \textbf{HERMES} system acts as a swift messenger between the initial prompt and the final, refined medical image segmentation.}
\end{quote}
In this case, \emph{HERMES} is the name of a framework, introduced through a creative but semantically correct analogy, yet incorrectly flagged as hallucinated. Conversely, false negatives occurred when the retrieval step returned semantically adjacent but non-supportive contexts. Another example from the same story fails to correctly assign the acronym \emph{FM} to \emph{Foundation Models}, resulting in the following undetected hallucination:
\begin{quote}
\emph{The proposed solution is an automated framework designed to enhance the accuracy of \textbf{flash memory (FM)-based segmentation}.}
\end{quote}

Because these two error types move in opposite directions, tuning the threshold did not lead to a stable operating point, making HHD unsuitable as a component for StoryScore.

\begin{table}[bt]
\caption{Comparison of explored hallucination detection methods.}
\label{tab:hallucination-methods}

\centering
\small
\rowcolors{2}{gray!10}{white}
\begin{tabularx}{\textwidth}{
    l
    >{\raggedright\arraybackslash}X
    >{\raggedright\arraybackslash}X
    >{\raggedright\arraybackslash}X
}
\toprule
\rowcolor{white}
\textbf{Method} & \textbf{What it detects} & \textbf{Strengths} & \textbf{Weaknesses} \\
\midrule
Capitalised Words       & Surface-form mismatch                  & Simple, transparent            & Noisy; overflags creativity \\
SpaCy NER               & Incorrect entity mentions              & Good precision                 & Misses conceptual hallucinations \\
MIRAGE                  & Rewrite instability                    & Captures semantic drift        & Penalises analogies \\
LLM-as-Judge (Qwen) & Factual consistency                    & Understands paraphrases        & Inconsistent; invents hallucinations \\
LLM-as-Judge (GPT~5.1)  & High-level reasoning errors            & Close to human judgement       & Too strict for storytelling \\
HHD              & Entity + retrieval alignment           & Balanced approach              & Unstable thresholds; mixed reliability \\
\end{tabularx}
\end{table}

Table~\ref{tab:hallucination-methods} summarizes our findings on different hallucination detection techniques, that lead to two key outcomes.
First, The most operationally stable detector was the simplest one, namely the NER-Based detection, combined with regex normalisation. In particular, it was the only method that:
\begin{enumerate*}
    \item remained stable across papers, showing consistent behaviour across different documents,
    \item did not penalise metaphors,
    \item detected genuine fabricated entities, and
    \item integrated cleanly with a software pipeline, without the need of external and costly API calls.
\end{enumerate*}
Second, being either too naive (NER), too strict (MIRAGE), or too unstable (LLM judge), the hallucination metric must have a reduced weight in the overall metric (limited to 10\%) to inform without overshadowing the other reliable qualities.

\section{Preliminary Findings}
\label{sec:findings}

We apply StoryScore to a set of 76 stories generated by our use case pipeline.
Table~\ref{tab:evaluation} reports aggregated statistics comparing two versions of the pipeline, using either a pre-trained version of Qwen2.5, or a fine-tuned version of the model. Fine-tuning substantially improves StoryScore and completely eliminates prompt leakage. Prompt Cleanliness and Title Coverage saturated, but their presence in StoryScore is a diagnostic safeguards.
Nevertheless, the pre-trained model still attains positive scores, particularly on BERTScore, Title Coverage, No Redundancy, and No Hallucination.

\begin{table}[tb]
\caption{StoryScore components on the test set, comparing the pre-trained and fine-tuned versions of the pipeline.}
\label{tab:evaluation}
\centering
\small
\begin{tabular}{lccccccc}
\textbf{LLM version} &
\rotatebox{90}{\textbf{StoryScore}} &
\rotatebox{90}{BERTScore} &
\rotatebox{90}{\makecell{Context\\Recall}} &
\rotatebox{90}{\makecell{Prompt\\Cleanliness}} &
\rotatebox{90}{\makecell{Title\\Coverage}} &
\rotatebox{90}{\makecell{No\\Redundancy}} &
\rotatebox{90}{\makecell{No\\Hallucination}} \\
\midrule
Pre-trained & 0.560 & 0.780 & 0.390 & 0.011 & 0.990 & 0.893 & 0.957 \\
Fine-tuned  & 0.787 & 0.815 & 0.472 & 1.000 & 0.998 & 0.903 & 0.925 \\
\bottomrule
\end{tabular}
\end{table}

A manual inspection found that narratives generated in the pre-trained settings are consistently affected by prompt leakage, redundancy, and excessive narrative filler, deficiencies that are only weakly penalized by global semantic metrics. The latters are tolerant of paraphrasing, repetition, and generic formulations, rewarding long and semantically “safe” texts despite being poorly readable. Conversely, the fine-tuned model generates more controlled, dense, and readable narratives, yet these qualitative improvements are only partially reflected in metrics.
Overall, variations in StoryScore were consistent with qualitative judgments of readability, narrative control, and factual grounding.

\section{Conclusions and Future Work}
\label{sec:conclusion}
We introduced StoryScore, a composite metric that integrates semantic alignment, lexical grounding, structural fidelity, fluency, and hallucination control, and analysed its behaviour on a concrete scientific storytelling use case.
The combination of complementary metrics provides a more informative picture of system behaviour than any single score alone. The initial evaluation suggests that StoryScore is useful for comparisons between generated stories, even if insufficient for discriminating critical aspects of narrative quality and text control in an absolute way. 

Hallucination detection proved to be an intrinsically complex task, even more when creativity and pedagogical reformulation are involved. Existing hallucination detection approaches are either too shallow, too rigid, or too unstable. Future evaluation frameworks should explicitly account for persona adaptation and narrative transformation, calling for a new definition of hallucination that distinguish between acceptable abstraction and factual distortion. Finally, this work motivates further refinement of composite metrics such as StoryScore -- in particular to address narrative control and structural validity -- as well as an assessment of alignment with human judgement following some best practices~\cite{chhun-etal-2022-human}.

\begin{acknowledgments}
This work was supported by the French Public Investment Bank (Bpifrance) i-Demo program within the LettRAGraph project (Grant ID DOS0256163/00).
\end{acknowledgments}

\section*{Declaration on Generative AI}
During the preparation of this work, the authors used GPT5.2 for grammar and spelling check. After using these tools, the authors reviewed and edited the content as needed and take full responsibility for the publication’s content.

\bibliography{bibliography}

@BOOK{national2017communicating,
  author    = {{National Academies of Sciences, Engineering, and Medicine}},
  title     = {{Communicating Science Effectively: A Research Agenda}},
  isbn      = "978-0-309-45102-4",
  doi       = "10.17226/23674",
  year      = 2017,
  publisher = "The National Academies Press",
  address   = "Washington, DC, USA"
}

@article{ji2023survey,
    author = {Ji, Ziwei and Lee, Nayeon and Frieske, Rita and Yu, Tiezheng and Su, Dan and Xu, Yan and Ishii, Etsuko and Bang, Ye Jin and Madotto, Andrea and Fung, Pascale},
    title = {{Survey of Hallucination in Natural Language Generation}},
    year = {2023},
    issue_date = {December 2023},
    publisher = {Association for Computing Machinery},
    address = {New York, NY, USA},
    volume = {55},
    number = {12},
    issn = {0360-0300},
    doi = {10.1145/3571730},
    journal = {ACM Comput. Surv.},
    month = mar,
    articleno = {248},
    numpages = {38}
}

@article{capili2024methods,
    author={Capili, Bernadette and Anastasi, Joyce K},
    journal={AJN The American Journal of Nursing},
    volume={124},
    number={7},
    pages={36--39},
    year={2024},
    publisher={LWW},
	isbn = {0002-936X},
	title = {{Methods to Disseminate Nursing Research: A Brief Overview}},
	doi = {10.1097/01.NAJ.0001025644.87717.4c}
}

@conference{afzal2023challenges,
    author={Anum Afzal and Juraj Vladika and Daniel Braun and Florian Matthes},
    title={{Challenges in Domain-Specific Abstractive Summarization and How to Overcome Them}},
    booktitle={15th International Conference on Agents and Artificial Intelligence - Volume 3: ICAART},
    year={2023},
    pages={682-689},
    publisher={SciTePress},
    organization={INSTICC},
    doi={10.5220/0011744500003393},
    isbn={978-989-758-623-1},
    issn={2184-433X},
}

@inproceedings{teleki-etal-2025-survey,
    title = {{A Survey on LLMs for Story Generation}},
    author = "Teleki, Maria  and
      Bengali, Vedangi  and
      Dong, Xiangjue  and
      Janjur, Sai Tejas  and
      Liu, Haoran  and
      Liu, Tian  and
      Wang, Cong  and
      Liu, Ting  and
      Zhang, Yin  and
      Shipman, Frank  and
      Caverlee, James",
    editor = "Christodoulopoulos, Christos  and
      Chakraborty, Tanmoy  and
      Rose, Carolyn  and
      Peng, Violet",
    booktitle = "Findings of the Association for Computational Linguistics: EMNLP 2025",
    month = nov,
    year = "2025",
    address = "Suzhou, China",
    publisher = "Association for Computational Linguistics",
    doi = "10.18653/v1/2025.findings-emnlp.750",
    pages = "13954--13966",
    ISBN = "979-8-89176-335-7"
}

@inproceedings{li2025generation,
    title = {{From Generation to Judgment: Opportunities and Challenges of LLM-as-a-judge}},
    author = {Li, Dawei  and
      Jiang, Bohan  and
      Huang, Liangjie  and
      Beigi, Alimohammad  and
      Zhao, Chengshuai  and
      Tan, Zhen  and
      Bhattacharjee, Amrita  and
      Jiang, Yuxuan  and
      Chen, Canyu  and
      Wu, Tianhao  and
      Shu, Kai  and
      Cheng, Lu  and
      Liu, Huan},
    booktitle = "2025 Conference on Empirical Methods in Natural Language Processing",
    month = nov,
    year = "2025",
    address = "Suzhou, China",
    publisher = "Association for Computational Linguistics",
    doi = "10.18653/v1/2025.emnlp-main.138",
    pages = "2757--2791",
    ISBN = "979-8-89176-332-6",
}

@misc{qwen2025qwen25technicalreport,
      title={{Qwen2.5 Technical Report}}, 
      author={Qwen and An Yang and Baosong Yang and Beichen Zhang and Binyuan Hui and Bo Zheng and Bowen Yu and Chengyuan Li and Dayiheng Liu and Fei Huang and Haoran Wei and Huan Lin and Jian Yang and Jianhong Tu and Jianwei Zhang and Jianxin Yang and Jiaxi Yang and Jingren Zhou and Junyang Lin and Kai Dang and Keming Lu and Keqin Bao and Kexin Yang and Le Yu and Mei Li and Mingfeng Xue and Pei Zhang and Qin Zhu and Rui Men and Runji Lin and Tianhao Li and Tianyi Tang and Tingyu Xia and Xingzhang Ren and Xuancheng Ren and Yang Fan and Yang Su and Yichang Zhang and Yu Wan and Yuqiong Liu and Zeyu Cui and Zhenru Zhang and Zihan Qiu},
      year={2025},
      eprint={2412.15115},
      archivePrefix={arXiv},
      primaryClass={cs.CL}
}

@misc{zhang2019bertscore,
      title={{BERTScore: Evaluating Text Generation with BERT}}, 
      author={Tianyi Zhang and Varsha Kishore and Felix Wu and Kilian Q. Weinberger and Yoav Artzi},
      year={2020},
      eprint={1904.09675},
      archivePrefix={arXiv},
      primaryClass={cs.CL}
}

@misc{spacy2020,
  author       = {Ines Montani and
                  Matthew Honnibal and
                  Adriane Boyd and
                  Sofie Van Landeghem and
                  Henning Peters},
  title        = {explosion/spaCy: v3.7.2: Fixes for APIs and
                   requirements
                  },
  month        = oct,
  year         = 2023,
  publisher    = {Zenodo},
  version      = {v3.7.2},
  doi          = {10.5281/zenodo.10009823},
}

@inproceedings{10.1145/3746252.3761644,
    author = {Vendeville, Benjamin and Ermakova, Liana and De Loor, Pierre and Kamps, Jaap},
    title = {{MIRAGE: A Metrics lIbrary for Rating hAllucinations in Generated tExt}},
    year = {2025},
    isbn = {9798400720406},
    publisher = {Association for Computing Machinery},
    address = {New York, NY, USA},
    doi = {10.1145/3746252.3761644},
    booktitle = {34th ACM International Conference on Information and Knowledge Management},
    pages = {6539–6543},
    numpages = {5},
    location = {Seoul, Republic of Korea},
    series = {CIKM '25}
}

@misc{chatgpt5.1,
  author       = {OpenAI},
  title        = {{ChatGPT 5.1}},
  year         = {2025},
  howpublished = {\url{https://chat.openai.com}},
  note         = {Large Language Model}
}

@article{roschelle2025generative,
  title={{Generative AI Unlocks the Power of Interactive Storytelling for Science Teachers and Learners}},
  author={Roschelle, Jeremy and Bansal, Mohit and Biswas, Gautam and Hmelo-Silver, Cindy and Lester, James},
  journal={Social Innovations Journal},
  volume={30},
  year={2025},
  url={https://socialinnovationsjournal.com/index.php/sij/article/view/9993}
}

@article{LI2023101311,
title = {{Teaching Data Science through Storytelling: Improving Undergraduate Data Literacy}},
journal = {Thinking Skills and Creativity},
volume = {48},
pages = {101311},
year = {2023},
issn = {1871-1871},
doi = {10.1016/j.tsc.2023.101311},
author = {You Li and Ye Wang and Yugyung Lee and Huan Chen and Alexis Nicolle Petri and Teryn Cha}
}

@inproceedings{wang202minilm,
author = {Wang, Wenhui and Wei, Furu and Dong, Li and Bao, Hangbo and Yang, Nan and Zhou, Ming},
title = {{MINILM: Deep Self-Attention Distillation for
Task-Agnostic Compression of Pre-Trained
Transformers}},
year = {2020},
isbn = {9781713829546},
publisher = {Curran Associates Inc.},
address = {Red Hook, NY, USA},
booktitle = {34th International Conference on Neural Information Processing Systems},
articleno = {485},
numpages = {13},
pages = {5776--5788},
location = {Vancouver, BC, Canada},
series = {NIPS '20}
}

@article{dahlstrom2014science_narrative,
author = {Dahlstrom, Michael F.},
title = {Using narratives and storytelling to communicate science with nonexpert audiences},
journal = {National Academy of Sciences},
volume = {111},
number = {supplement\_4},
pages = {13614-13620},
year = {2014},
doi = {10.1073/pnas.1320645111}}

@article{freeman2024recall,
    author = {Freeman, Alexandra L. J. and Tanase, Lisa-Maria and Schneider, Claudia R. and Kerr, John},
    title = {Can narrative help people engage with and understand information without being persuasive? An empirical study},
    journal = {Royal Society Open Science},
    volume = {11},
    number = {7},
    pages = {231708},
    year = {2024},
    month = {07},
    issn = {2054-5703},
    doi = {10.1098/rsos.231708}
}

@inproceedings{zhang2025revtogether,
author = {Zhang, Yu and Fu, Kexue and Lu, Zhicong},
title = {{RevTogether: Supporting Science Story Revision with Multiple AI Agents}},
year = {2025},
isbn = {9798400713958},
publisher = {Association for Computing Machinery},
address = {New York, NY, USA},
doi = {10.1145/3706599.3719888},
booktitle = {Extended Abstracts of the CHI Conference on Human Factors in Computing Systems},
articleno = {462},
numpages = {7},
pages={1--7},
location = {Yokohama, Japan},
series = {CHI EA '25}
}

@inproceedings{lin-2004-rouge,
    title = {{ROUGE: A Package for Automatic Evaluation of Summaries}},
    author = "Lin, Chin-Yew",
    booktitle = "Text Summarization Branches Out",
    month = jul,
    year = "2004",
    address = "Barcelona, Spain",
    publisher = "Association for Computational Linguistics",
    url = "https://aclanthology.org/W04-1013/",
    pages = "74--81"
}

@inproceedings{scialom-etal-2021-questeval,
    title = {{QuestEval: Summarization Asks for Fact-based Evaluation}},
    author = "Scialom, Thomas  and
      Dray, Paul-Alexis  and
      Lamprier, Sylvain  and
      Piwowarski, Benjamin  and
      Staiano, Jacopo  and
      Wang, Alex  and
      Gallinari, Patrick",
    editor = "Moens, Marie-Francine  and
      Huang, Xuanjing  and
      Specia, Lucia  and
      Yih, Scott Wen-tau",
    booktitle = "2021 Conference on Empirical Methods in Natural Language Processing",
    month = nov,
    year = "2021",
    address = "Online and Punta Cana, Dominican Republic",
    publisher = "Association for Computational Linguistics",
    doi = "10.18653/v1/2021.emnlp-main.529",
    pages = "6594--6604"
}

@inproceedings{zhao-etal-2019-moverscore,
    title = {{MoverScore: Text Generation Evaluating with Contextualized Embeddings and Earth Mover Distance}},
    author = "Zhao, Wei  and
      Peyrard, Maxime  and
      Liu, Fei  and
      Gao, Yang  and
      Meyer, Christian M.  and
      Eger, Steffen",
    editor = "Inui, Kentaro  and
      Jiang, Jing  and
      Ng, Vincent  and
      Wan, Xiaojun",
    booktitle = "2019 Conference on Empirical Methods in Natural Language Processing and the 9th International Joint Conference on Natural Language Processing (EMNLP-IJCNLP)",
    month = nov,
    year = "2019",
    address = "Hong Kong, China",
    publisher = "Association for Computational Linguistics",
    doi = "10.18653/v1/D19-1053",
    pages = "563--578"
}

@inproceedings{wang-etal-2020-asking,
    title = {{Asking and Answering Questions to Evaluate the Factual Consistency of Summaries}},
    author = "Wang, Alex  and
      Cho, Kyunghyun  and
      Lewis, Mike",
    editor = "Jurafsky, Dan  and
      Chai, Joyce  and
      Schluter, Natalie  and
      Tetreault, Joel",
    booktitle = "58th Annual Meeting of the Association for Computational Linguistics",
    month = jul,
    year = "2020",
    address = "Online",
    publisher = "Association for Computational Linguistics",
    doi = "10.18653/v1/2020.acl-main.450",
    pages = "5008--5020"
}

@inproceedings{chhun-etal-2022-human,
    title = {{Of Human Criteria and Automatic Metrics: A Benchmark of the Evaluation of Story Generation}},
    author = "Chhun, Cyril  and
      Colombo, Pierre  and
      Suchanek, Fabian M.  and
      Clavel, Chlo{\'e}",
    booktitle = "29th International Conference on Computational Linguistics",
    month = oct,
    year = "2022",
    address = "Gyeongju, Republic of Korea",
    publisher = "International Committee on Computational Linguistics",
    url = "https://aclanthology.org/2022.coling-1.509/",
    pages = "5794--5836"
}

@article{huang2025survey,
author = {Huang, Lei and Yu, Weijiang and Ma, Weitao and Zhong, Weihong and Feng, Zhangyin and Wang, Haotian and Chen, Qianglong and Peng, Weihua and Feng, Xiaocheng and Qin, Bing and Liu, Ting},
title = {{A Survey on Hallucination in Large Language Models: Principles, Taxonomy, Challenges, and Open Questions}},
year = {2025},
issue_date = {March 2025},
publisher = {Association for Computing Machinery},
address = {New York, NY, USA},
volume = {43},
number = {2},
issn = {1046-8188},
doi = {10.1145/3703155},
journal = {ACM Trans. Inf. Syst.},
month = jan,
articleno = {42},
numpages = {55}
}

@inproceedings{janiak-etal-2025-illusion,
    title = {{The Illusion of Progress: Re-evaluating Hallucination Detection in LLMs}},
    author = "Janiak, Denis  and
      Binkowski, Jakub  and
      Sawczyn, Albert  and
      Gabrys, Bogdan  and
      Shwartz-Ziv, Ravid  and
      Kajdanowicz, Tomasz Jan",
    editor = "Christodoulopoulos, Christos  and
      Chakraborty, Tanmoy  and
      Rose, Carolyn  and
      Peng, Violet",
    booktitle = "2025 Conference on Empirical Methods in Natural Language Processing",
    month = nov,
    year = "2025",
    address = "Suzhou, China",
    publisher = "Association for Computational Linguistics",
    doi = "10.18653/v1/2025.emnlp-main.1761",
    pages = "34728--34745",
    ISBN = "979-8-89176-332-6"
}

@inproceedings{gao2024training,
  title={{Training like a medical resident: Context-prior learning toward universal medical image segmentation}},
  author={Gao, Yunhe},
  booktitle={IEEE/CVF Conference on Computer Vision and Pattern Recognition},
  pages={11194--11204},
  year={2024}
}

@inproceedings{sillano2026personas,
    author = {Sillano, Andrea and {De Russis}, Luigi and Cal\'o, Tommaso and Troncy, Raphael and Lisena, Pasquale},
    title = {{Mapping Personas to Text Transformations: A Taxonomy Outline for Content Adaptation}},
    booktitle = {From Generation to Simulation: Responsible Use of AI Personas in Human-Centered Design and Research (ACM CHI Workshop)},
    year = {2026},
    publisher={CEUR-WS},
}


\clearpage
\end{document}